\documentclass[runningheads]{llncs}
\usepackage[T1]{fontenc}
\usepackage{graphicx}
\usepackage{lipsum}
\usepackage{multirow}
\usepackage{adjustbox}
\usepackage{xcolor}
\usepackage{amsmath}
\begin{document}
%
%\title{A Block-Based Approach \\to Discriminate Synthetic from Real Images}
\title{DeepFeatureX Net: Deep Features eXtractors based Network for discriminating synthetic from real images}
\author{Orazio Pontorno\orcidID{0009-0009-0381-9971} \and
Luca Guarnera\orcidID{0000-0001-8315-351X} \and
Sebastiano Battiato\orcidID{0000-0001-6127-2470}}
\authorrunning{O. Pontorno et al.}
\institute{Department of Mathematics and Computer Science, University of Catania, Italy\\
\email{orazio.pontorno@phd.unict.it}, \email{\{luca.guarnera,sebastiano.battiato\}@unict.it}}
\maketitle              % typeset the header of the contribution
\begin{abstract}
% Deepfakes, synthetic images generated by deep learning algorithms, represent one of the greatest challenges in the field of Digital Forensics. In the scientific community, we are working to develop methodologies that can discriminate the nature of digital images. However, these methodologies face the challenge of generalization, which is the ability to recognize the nature of an image even if generated by an architecture not seen during training. In this context we propose an innovative approach based on three blocks called Base Models, each of which is responsible for extracting the discriminative features of a specific category of images (generated by diffusion Models, generated by GAN or real). The results show that this approach is actually more performant in generalization than the use of standard models.

Deepfakes, synthetic images generated by deep learning algorithms, represent one of the biggest challenges in the field of Digital Forensics. The scientific community is working to develop approaches that can discriminate the origin of digital images (real or AI-generated). However, these methodologies face the challenge of generalization, that is, the ability to discern the nature of an image even if it is generated by an architecture not seen during training. This usually leads to a drop in performance. In this context, we propose a novel approach based on three blocks called Base Models, each of which is responsible for extracting the discriminative features of a specific image class (Diffusion Model-generated, GAN-generated, or real) as it is trained by exploiting deliberately unbalanced datasets. The features extracted from each block are then concatenated and processed to discriminate the origin of the input image. Experimental results showed that this approach not only demonstrates good robust capabilities to JPEG compression but also outperforms state-of-the-art methods in several generalization tests. Code, models and dataset are available at \url{https://github.com/opontorno/block-based_deepfake-detection}.

\end{abstract}
\section{Introduction}
Generative models have achieved a high degree of fidelity in content generation, producing increasingly realistic and convincing results. Thanks to the vast amount of data available today and the continuous development of complex architectures, such as Generative Adversarial Networks (GANs)~\cite{goodfellow2014generative} and Diffusion Models (DMs)~\cite{ho2020denoising,sohl2015deep}, these models are able to produce images, text, sound and video with an astonishing quality that can hardly be distinguished from those created by human beings. This ability to generate high-fidelity content has opened up new opportunities in a wide range of fields, from art and entertainment to scientific research and multimedia content production. However, along with their powerful creative capabilities, generative models also have several negative aspects. One of the main problems is the possibility of abuse, as such models can be used to generate fake or convincingly manipulated content, fuelling the spread of misinformation and fraud~\cite{shan2023glaze,vyas2023provable}. Moreover, they can raise ethical concerns regarding intellectual property and privacy~\cite{leotta2023not}, especially when they are used to create content based on personal data without the consent of the involved people. The proper and preventive detection of AI-generated content therefore becomes a critical priority to combat the spread of deepfakes and maintain the integrity of online information.

The scientific community is striving to find increasingly new and effective techniques and methods that can discern the nature (real or generated) of digital images. These techniques can be based on analysis and processing of statistics extracted from images (e.g. analytical traces) or on deep learning engines. 
% Of the first category are all those techniques based on 
Among other we recall the analysis of image frequencies, 
%the most common techniques use mathematical transforms 
such as the Discrete Cosine Transform (DCT) and the Fourier Transform to map image pixels from the spatial domain to the frequency domain,
%The use of these allows one to go deep into the search for discriminative traces contained in domains other than the spatial domain, 
facilitating greater interpretability in the task of deepfake recognition~\cite{asnani2023reverse,guarnera2022exploitation}. Deep learning-based methodologies involve the construction of neural models 
%capable of solving the classification task. In general, they 
achieving in general better results than the previous techniques~\cite{arshed2024multiclass,gandhi2020adversarial}, but at the expense of a lower generalization. %interpretability and leak generalization. 

In this paper we propose a deep learning based architecture that exploits three backbones, called ``Base Models'' (BM) %(see Section~\ref{sec:method} for more details), each of which has the task of extracting the discriminative features of each specific class of images (real, GAN-generated, and DM-generated).
trained and specialized to specific classification tasks with special emphasis to DM generated data, GAN generated data, and real ones. 
% \begin{itemize}
%     \item BM$_1$ - learned to detect mainly Diffusion Models (DMs) generated data.
%     \item BM$_2$ - learned to detect mainly Generative Adversarial Networks (GANs) generated data.
%     \item BM$_3$ - learned to detect mainly real data.
% \end{itemize}
% Given as input a generic image $\mathcal{I}$ and representing the three BMs as functions $\phi$, 3 feature vectors $\phi_c(\mathcal{I}) \quad \forall c \in \mathcal{C}=\{ \textit{DMs, GANs, reals} \}$ are obtained from BMs and their concatenation $\phi(\mathcal{I})=\phi_{\small{DM}}(\mathcal{I})\oplus\phi_{\small{GAN}}(\mathcal{I})\oplus\phi_{\small{REAL}}(\mathcal{I})$ is processed. 
The fundamental concept is based on utilising the inherent capabilities of the basic models, each of which is dedicated to extracting discriminating features specific to a generating architecture left behind during the image generation process. This approach aims at enhancing the final model by making it more resilient and robust to JPEG compression attacks, commonly employed by social networks, and more effective in the generalisation of the acquired knowledge. Focusing on specific distinctive features associated with different image generation technologies allows the model to develop a deeper and more focused understanding of the peculiarities of each image category, thus improving its ability to distinguish between genuine and synthetic images in real and variable contexts.
%\textbf{Motivation.} What prompted us to undertake this research is
With this work we face the difficulty, often encountered in the state-of-the-art, of generalizing the recognition capabilities acquired in the training phase both to images generated by AIs not belonging to the dataset used in that phase and to synthetic images of generating architectures other than those taken into consideration.\\
The main contributions of this paper are:
\begin{itemize}
    \item A new approach for extracting main features from digital images using Base Models.
    \item A model capable of retaining its discriminative ability even at JPEG compression attacks.
\end{itemize}

The paper follows the following structure: Section \ref{sec:sota} provides an overview of the main deepfake detection methods currently present in the state-of-the-art; in Section \ref{sec:dataset}, a detailed description of the dataset of images used to conduct the experiments is provided; subsequently, in Section \ref{sec:method}, the architecture proposed in this study and the stages of the training method are presented in detail; the experimental results obtained during the testing phase are reported in a Section \ref{sec:results}; finally, the paper concludes with a concluding section where the main findings are summarized and the future directions of research are outlined.
\label{sec:intro}

\section{Related Works}
Most deepfake detection methods are based on intrinsic trace analysis to classify real content from synthetic ones. 
The Expectation-Maximization algorithm was used in~\cite{guarnera2020fighting} to capture the correlation between pixels, resulting in a discriminative trace able to distinguish deepfake images from pristine ones. McCloskey et al.~\cite{mccloskey2019detecting} showed that generative models create synthetic content with color channel curve statistics different from the real data, resulting in another discriminative trace. In the frequency domain~\cite{guarnera,Marra2019DoGL}, researchers highlighted the possibility of identifying abnormal traces left during generative models, mainly analyzing features extracted from DCT~\cite{bergmann2024forensic,concas2022tensor,JOEL_ICML_2020}. 
Liu et al.~\cite{liu2021spatial} proposed a method called Spatial-Phase Shallow Learning (SPSL) that combines spatial imaging and phase spectrum to capture artifacts from up-sampling on synthetic data, improving deepfake detection. Corvi et al.~\cite{corvi2023intriguing} analyzed a large number of images generated by different families of generative models (GAN, DM, and VQ-GAN (Vector Quantized Generative Adversarial Networks)) in the Fourier domain to discover the most discriminative features between real and synthetic images. The experiments showed that regular anomalous patterns are available in each category of involved architecture. 
Another category of detectors are deep neural network-based approaches. Wang et al.~\cite{wang2020cnn} used a ResNet-50 model trained with images generated by ProGAN~\cite{karras2018progressive} to differentiate real from synthesized images. Their study demonstrated the model's ability to generalize beyond ProGAN-generated Deepfakes. Wang et al.~\cite{wang2021fakespotter} introduced FakeSpotter, a new approach that relies on monitoring the behaviors of neurons (counting which and how many activate on the input image) within a dedicated CNN to identify Deepfake-generated faces. 
Many researchers have focused their research on trying to investigate how possible it is to detect images created by diffusion models. Corvi et al. \cite{corvi2023detection} were among the first to address this issue, exploring the difficulties in distinguishing images generated by diffusion models from real ones and evaluating the suitability of current detectors.
Sha et al. \cite{sha2023fake} proposed DE-FAKE, a machine learning classifier designed for detecting diffusion model-generated images across four prominent text-to-image architectures. 
The authors then proposed a pioneering study on the detection and attribution of fake images generated by diffusion models, demonstrating the feasibility of distinguishing such images from real ones and attributing them to the source models, and also discovering the influence of prompts on the authenticity of images.
Recently, Guarnera et al.\cite{guarnera2024mastering} proposed a method based on the attribution of images generated by generative adversarial networks (GANs) and diffusion models (DMs) through a multi-level hierarchical strategy. At each level, a distinct and specific task is addressed: the first level (more generic), allows discerning between real and AI-generated images (either created by GAN or DM architectures); the second level determines whether the images come from GAN or DM technologies; and the third level addresses the attribution of the specific model used to generate the images.

The limitations of these methods mainly concern the presence of experimental results performed only under ideal conditions and, consequently, the almost total absence of generalization tests: the classification performance of most state-of-the-art methods drops drastically when testing images generated by architectures never considered during the training procedure. 

\label{sec:sota}

\section{Dataset details}
The dataset comprises a total of $72,334$ images, distributed as follows: $19,334$ real images collected from CelebA~\cite{liu2015faceattributes}, FFHQ~\cite{karras2019style}, and other sources~\cite{leotta2023not,corvi2023detection}, $37,572$ images generated by the GAN architectures GauGAN~\cite{park2019gaugan}, BigGAN~\cite{brock2018large}, ProGAN~\cite{karras2017progressive}, StarGAN~\cite{choi2018stargan}, AttGAN~\cite{he2019attgan}, GDWCT~\cite{cho2019image}, CycleGAN~\cite{zhu2017unpaired}, StyleGAN~\cite{karras2019style}, StyleGAN2~\cite{karras2020analyzing}, StyleGAN3~\cite{karras2021alias}, and $15,423$ images produced by the DM architectures DALL-E MINI~\footnote{\href{https://github.com/borisdayma/dalle-mini.}{github.com/borisdayma/dalle-mini}}, DALL-E 2~\cite{ramesh2022hierarchical}, Latent Diffusion~\cite{rombach2022high}, Stable Diffusion~\footnote{\href{https://github.com/CompVis/stable-diffusion}{github.com/CompVis/stable-diffusion}} (Figure~\ref{fig:pipeline} (a) shows some examples of used images). All images are in PNG format.

Initially the dataset was divided into three parts: a first $40\%$ was used for training and validation of the Base Models (refer to Section~\ref{sub:base_models}); another $40\%$ was used for training and validation of the complete models (refer to Section~\ref{sub:overall}); finally the remaining $20\%$ was used as testing dataset for both phases. 
% Highlighting the research goals, the diversity of the dataset took precedence. As a result, images were selected to reduce the impact of their meaning, concentrating solely on the fundamental generative structure. This method guarantees that the dataset reflects a range of image generation methods, irrespective of their specific subject matter. 
Since our only goal is to discern the nature of the images, regardless of semantics, resolution, and size, the images were collected with as much variety of these parameters as possible. 
The objective is to underscore the dataset's varied composition, incorporating images from different sources, each marked by unique tasks and approaches to image creation. %Table \ref{tab:dataset} offers a detailed breakdown of the dataset~\footnote{Please note that Stable Diffusion and DALL-E MINI were obtained from the links \href{https://github.com/CompVis/stable-diffusion}{github.com/CompVis/stable-diffusion} and \href{https://github.com/borisdayma/dalle-mini.}{github.com/borisdayma/dalle-mini}}, presenting both the total number of images for each architectural category and the diverse sources from which they were obtained.
\label{sec:dataset}

\section{Proposed Method}
\begin{figure}[t!]
    \centering
    \includegraphics[width=0.9\textwidth]{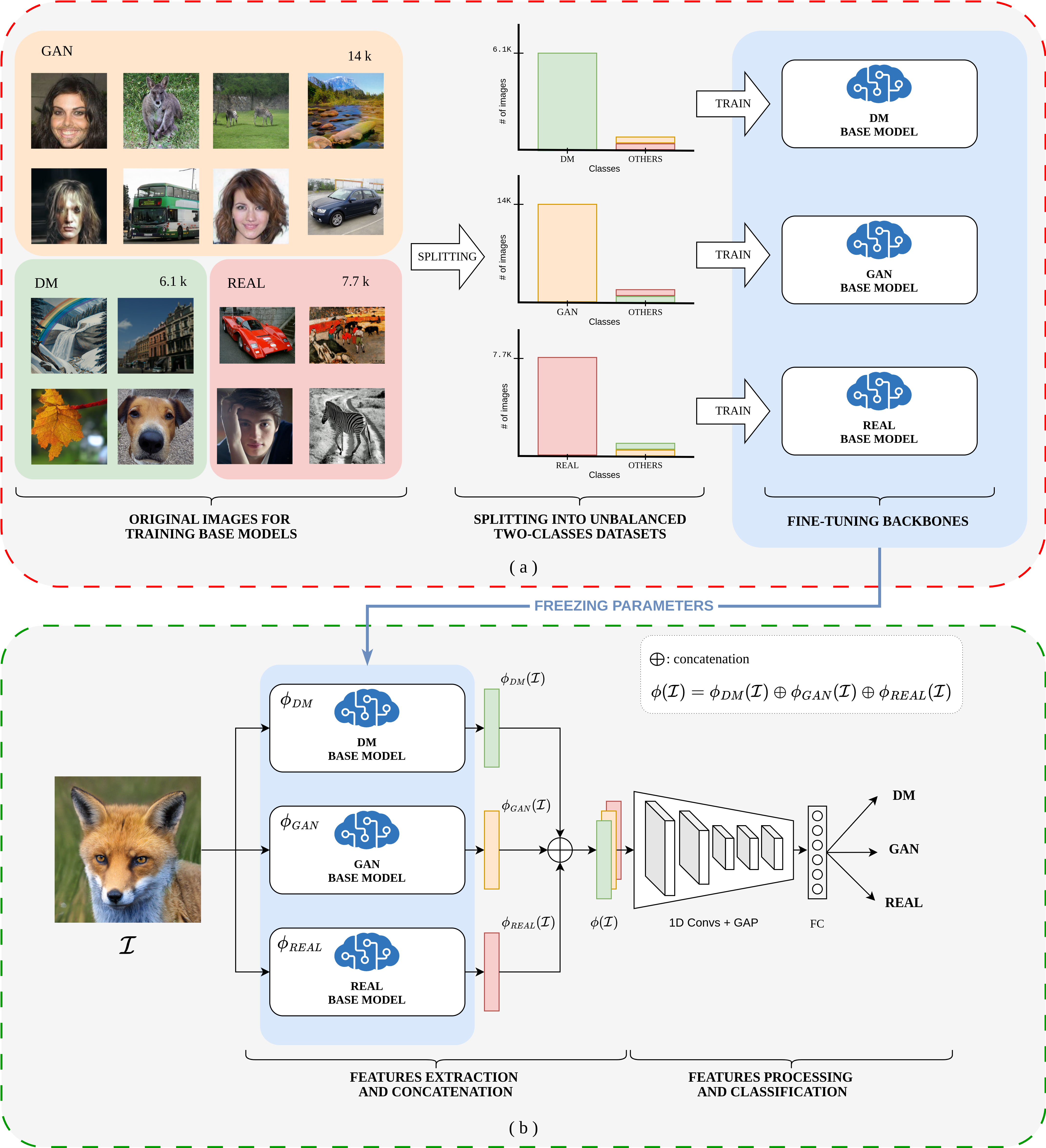}
    \caption{Entire pipeline of the proposed method. (a) shows the process of dividing the training dataset into three unbalanced subsets, each with respect to a specific class (DM, GAN, real) used for training a specific Base Model. (b) illustrates the architecture of the final model, which takes the three Base Models  $\phi_c$ trained in the previous phase with frozen weights, and uses them to extract the features from a digital image $\phi_c(\mathcal{I})$, where $c \in \mathcal{C}=\{\small{DM}, \small{GAN}, \small{REAL}\}$. These are then concatenated in channel dimension $\phi(\mathcal{I})=\phi_{\small{DM}}(\mathcal{I})\oplus\phi_{\small{GAN}}(\mathcal{I})\oplus\phi_{\small{REAL}}(\mathcal{I})$ and processed to solve the classification task.}
    \vspace{-0.4cm}
    \label{fig:pipeline}
\end{figure}

The model proposed in this paper consists of exploiting three CNN backbones as feature extractors, which are then concatenated and processed to solve the classification task. The key idea of the model lies in the training of the three backbones, each of which is trained using a specially unbalanced dataset of images (as detailed below). The purpose of this procedure is to force each backbone to focus on finding the discriminative features, left by each type of generative model during the generation phase, contained in the images belonging to a specific class (real, GAN-generated, DM-generated). We give the name of `Base Model' to backbones trained on a highly unbalanced dataset and later used as feature pullers in the complete model. Figure~\ref{fig:pipeline} shows the entire pipeline of the proposed method.

\subsection{Training of Base Models} \label{sub:base_models}
As mentioned above, each of the three Base Models was trained using a subset of the training data set. In particular, from the original image are extracted three subsets that are somewhat unbalanced with respect to each of the classes following a 90:10 ratio. In this training phase, a pre-trained Convolutional Neural Network (CNN) standard is adapted by performing a binary classification between the predominant class and the one named `others', composed of some images taken randomly by the other two remaining classes. Figure~\ref{fig:pipeline} (a) summarizes the overall process. 
Once the training is completed, the three Base Models are first frozen, the last linear layer (delegated to the binary classification) removed so that the characteristics maps of the last convolution layer are returned as output.
Our hypothesis, verified during the test phase, is that, following this training procedure, the backbones focus on the search for the main characteristics of the predominant class in order to be able to recognize their presence/absence, during inference. In conducting the experiments, the following CNNs were used as backbone: DenseNet 121, DenseNet 161, DenseNet 169, DenseNet 201~\cite{huang2017densely}, EfficientNet b0, EfficientNet b4~\cite{tan2019efficientnet}, ResNet 18, ResNet 34, ResNet 50, ResNet 101, ResNet 152~\cite{he2016deep}, ResNeXt 101~\cite{xie2017aggregated}, ViT b16, ViT b32~\cite{dosovitskiy2020image}. All backbones have been pretrained on the Imagenet~\cite{5206848} dataset. All experiments were conducted on GPU NVIDIA RTX a6000. The parameters of each model
were selected by choosing those that obtained the minimum loss value during model validation. \\
Table~\ref{tab:backbone_results} shows the accuracy, recall, precision, and F1 score values obtained by evaluating all backbones on testing images. From the results we can observe how this training led to maximizing the recall value, this indicates that the classification model is able to correctly identify all the positive examples of the interest class (the unbalanced one). In other words, the model tends to minimize false negatives; that is, there are no cases where the model wrongly classified a positive example as negative. This confirms our initial hypothesis that, following the training procedure described above, Base Models are able to capture the discriminative features of each generating architecture.

\begin{table}[t!]
    \centering
    \begin{adjustbox}{max width=1\textwidth}
    \setlength{\tabcolsep}{2pt}
    \begin{tabular}{|c|cccc|cccc|cccc|}
        \hline
        \multirow{2}{*}{\texttt{Backbone}} & \multicolumn{4}{c|}{DM Base Model}&\multicolumn{4}{c|}{GAN Base Model}&\multicolumn{4}{c|}{REAL Base Model} \\
         & \textbf{Acc} & \textbf{Rec} & \textbf{Pre} & \textbf{F1} & \textbf{Acc} & \textbf{Rec} & \textbf{Pre} & \textbf{F1} & \textbf{Acc} & \textbf{Rec} & \textbf{Pre} & \textbf{F1} \\
        \hline
        DenseNet 121 & 76.34 & 99.00 & 47.60 & 64.29 & 92.34 & 99.25 & 87.64 & 93.08 & 73.17 & 99.08 & 49.73 & 66.22 \\
        \hline
        DenseNet 161 & 83.96 & 98.64 & 57.40 & 72.57 & 94.62 & 99.45 & 91.03 & 95.05 & 77.39 & 99.21 & 54.04 & 69.97 \\
        \hline
        DenseNet 169 & 78.83 & 99.16 & 50.40 & 66.83 & 93.86 & 99.32 & 89.91 & 94.38 & 71.20 & 99.29 & 47.95 & 64.67 \\
        \hline
        DenseNet 201 & 79.11 & 98.90 & 50.75 & 67.08 & 92.61 & 99.37 & 87.96 & 93.32 & 72.97 & 99.29 & 49.54 & 66.10 \\
        \hline
        EfficientNet b0 & 85.74 & 97.09 & 60.50 & 74.55 & 88.77 & 98.81 & 82.87 & 90.14 & 77.91 & 97.56 & 54.70 & 70.10 \\
        \hline
        EfficientNet b4 & 78.00 & 97.47 & 49.43 & 65.59 & 87.14 & 98.74 & 80.78 & 88.86 & 74.31 & 97.27 & 50.84 & 66.78 \\
        \hline
        ResNet 18 & 76.64 & 98.03 & 47.90 & 64.36 & 84.28 & 99.06 & 77.16 & 86.75 & 65.14 & 99.03 & 43.17 & 60.13 \\
        \hline
        ResNet 34 & 77.29 & 98.12 & 48.62 & 65.02 & 83.33 & 99.40 & 75.93 & 86.10 & 66.75 & 98.82 & 44.33 & 61.21 \\
        \hline
        ResNet 50 & 78.14 & 98.61 & 49.59 & 66.00 & 90.34 & 99.14 & 84.82 & 91.42 & 70.79 & 99.13 & 47.59 & 64.31 \\
        \hline
        ResNet 101 & 77.20 & 99.00 & 48.53 & 65.13 & 90.85 & 98.87 & 85.71 & 91.82 & 69.10 & 99.08 & 46.17 & 62.99 \\
        \hline
        ResNet 152 & 76.48 & 99.00 & 47.75 & 64.43 & 93.42 & 99.32 & 89.23 & 94.00 & 70.59 & 98.92 & 47.41 & 64.10 \\
        \hline
        ResNeXt 101 & 75.38 & 98.58 & 46.60 & 63.28 & 93.82 & 98.40 & 90.53 & 94.30 & 66.26 & 98.66 & 43.96 & 60.82 \\
        \hline
        ViT b16 & 76.53 & 98.58 & 47.80 & 64.38 & 83.58 & 99.69 & 76.11 & 86.32 & 68.45 & 99.24 & 45.67 & 62.55 \\
        \hline
        ViT b32 & 74.05 & 96.92 & 45.20 & 61.65 & 80.83 & 98.99 & 73.39 & 84.29 & 60.44 & 99.61 & 40.13 & 57.21 \\
        \hline
    \end{tabular}
    \end{adjustbox}
    \caption{Percentage values of the metrics Accuracy, Recall, Precision, and F1 Score obtained by testing the Base Model to the binary classification between the predominant class and the class 'others'.}
    \label{tab:backbone_results}
    \vspace{-1cm}
\end{table}

\subsection{Overall architecture} \label{sub:overall}
The final model uses the three Base Models trained as described in Section~\ref{sub:base_models} as feature extractors, at this stage they will no longer be trained as the weights have been frozen. Each base model receives the same digital image as input and is tasked with identifying and extracting the discriminative features of each class. These are then concatenated to obtain a three-channel tensor, which is then processed through a custom CNN, consisting of a sequence of 5 convolutions 1D with respectively kernel size of 7, 5, 3, 3, 3, all with padding 1 and stride 1; This was followed by a Global Average Pooling operation and a three-node linear output classifier. Figure~\ref{fig:pipeline} (b) presents both the entire pipeline and a graphical representation of the model.

For the training phase of the complete models we used the Cross Entropy Loss weighed with respect to the frequency of each class in the dataset (Equation~\ref{eq:wcel}). This choice was necessary to avoid that the models were too influenced by the imbalance present in the dataset of used images. 

\begin{equation} \label{eq:wcel}
\text{Weighted Cross Entropy Loss} = -\frac{1}{N} \sum_{i=1}^{N} \sum_{c \in \mathcal{C}} w_c y_{i,c} \log(\hat{y}_{i,c})
\end{equation}

where $N$ s the number of samples, $\mathcal{C}=\{ \text{GAN}, \text{DM}, \text{REAL}\}$ is the set of classes, $y_{i,c}$ is the ground truth label for sample $i$ and $c$, $\hat{y}_{i,c}$ is the predicted probability for sample $i$ and class $c$, and $w_c$ is the weight for class $c$. In particular:

\begin{equation*}
    w_{\text{c}} = \frac{1}{\# \textit{images of class c}} \qquad \forall c \in \mathcal{C}.
\end{equation*}
\label{sec:method}

\section{Experimental results}
Two types of experiments were conducted: Inference and robustness tests to assess the effectiveness and robustness of the classification models, and comparison with the state-of-the-art in the generalization test.

\subsection{Inference and robustness tests} \label{sub:inf_rob}
In this first testing phase, we tested the proposed architecture by varying the backbone of the Base Model in order to choose the best model. For this testing phase, $20\%$ of the images of the original dataset (Sec.~\ref{sec:dataset}) were used. Furthermore, in order to make the accuracy metric more meaningful, both validation and testing datasets were balanced so as to have the same number of images for each class (DM-generated, GAN-generated, real).

\begin{figure}[t]
    \centering
    \includegraphics[width=0.8\textwidth]{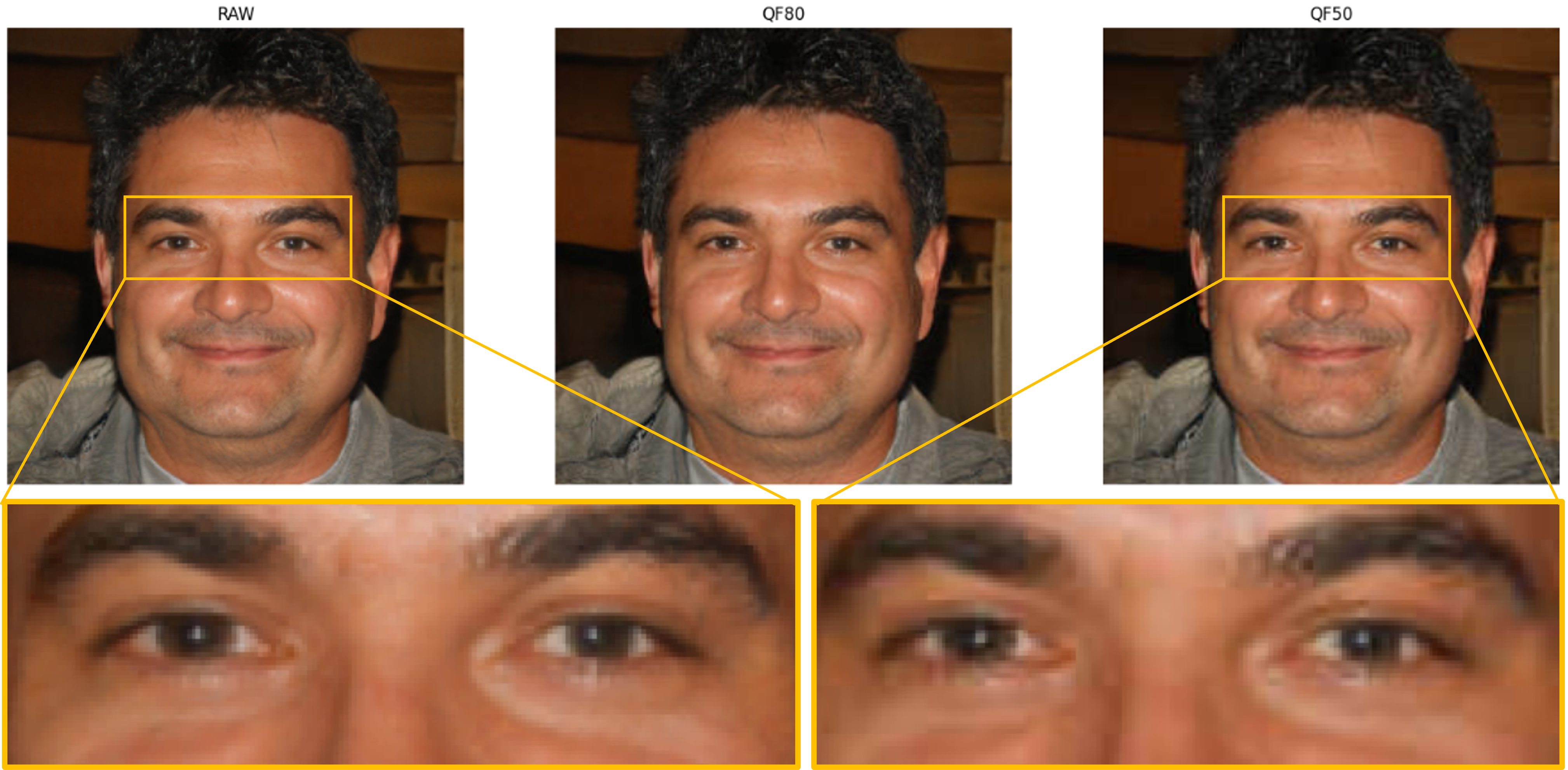}
    \caption{Image variation as JPEG compression Quality Factor decreases. On the left raw image, at center JPEG compressed image at Quality Factor 80, and on the right the image at QF 50. Image generated by StyleGAN2~\cite{karras2020analyzing}.}
    \label{fig:jpeg_wolf}
\end{figure}

Initially, the models were tested using just raw images. Subsequently, the test images were compressed into JPEG format using different Quality Factors (QF): 90, 80, 70, 60, 50. On these new compressed image sets, we again tested the models by analysing their robustness to JPEG compression.
Figure~\ref{fig:jpeg_wolf} shows the main differences between images with and without JPEG compression. It can be observed that as QF is decreased, low frequencies are removed and JPEG blocks are visible. This operation could lead to the removal of those (potentially) discriminative features identified by the various classifiers.

Table~\ref{tab:three-inner_testing} shows the performance of both tests in the three-class classification. From the results obtained, we can see that, regardless of the backbone used in the Base Model, in general this approach succeeds in achieving accuracy values in excess of $85\%$. In particular, the use of a model belonging to the DenseNet family as a backbone gives a boost to the overall performance of the models.

\begin{table}[t!]
    \centering
    \setlength{\tabcolsep}{4pt}
    \begin{adjustbox}{max width=1\textwidth}
    \begin{tabular}{|c|cccccc|}
        \hline
        \multirow{3}{*}{\begin{tabular}[c]{@{}c@{}}\texttt{Base Model} \\ \texttt{backbone}\end{tabular}} & \multicolumn{6}{c|}{\texttt{Multi-class  -  Accuracy / F1-Score ($\%$)}} \\
         & \multirow{2}{*}{\textit{RAW}} & \multicolumn{5}{c|}{\textit{JPEG-Compression}} \\ \cline{3-7}
         &  & QF90 & QF80 & QF70 & QF60 & QF50 \\
        \hline
        DenseNet 121 & 91.39/91.35 & 89.34/89.26 & 87.19/87.11 & 84.31/84.26 & 81.17/81.24 & 79.19/79.33 \\
        \hline
        DenseNet 161 & \textbf{93.30/93.21} & \textbf{91.72/92.07} & \textbf{88.21/88.04} & 83.88/84.06 & 77.78/78.21 & 74.06/74.53 \\
        \hline
        DenseNet 169 & 91.11/91.14 & 89.36/89.36 & 86.99/87.04 & 82.99/83.17 & 77.92/78.31 & 75.46/75.92 \\
        \hline
        DenseNet 201 & 92.31/92.30 & 90.70/90.67 & 87.78/87.76 & 84.76/84.76 & 81.98/82.01 & 79.78/79.87 \\
        \hline
        EfficientNet b0 & 88.62/88.66 & 86.19/86.26 & 84.74/84.83 & 82.95/83.05 & 80.45/80.56 & 78.98/79.11 \\
        \hline
        EfficientNet b4 & 84.39/84.51 & 83.00/83.16 & 81.59/81.75 & 80.28/80.45 & 78.48/78.66 & 78.42/78.61 \\
        \hline
        ResNet 18 & 85.22/85.23 & 84.64/84.64 & 83.61/83.61 & 82.72/82.71 & 80.65/80.68 & 79.77/79.77 \\
        \hline
        ResNet 34 & 85.98/86.00 & 85.37/85.37 & 84.25/84.23 & 82.82/82.75 & 80.37/80.30 & 79.12/79.02 \\
        \hline
        ResNet 50 & 89.63/89.62 & 88.01/87.99 & 86.07/86.06 & 83.94/83.95 & 80.90/80.96 & 78.85/78.93 \\
        \hline
        ResNet 101 & 90.96/90.97 & 89.85/89.84 & 87.90/87.88 &\textbf{86.53/86.52} & 83.27/84.32 & \textbf{82.97/83.05} \\
        \hline
        ResNet 152 & 91.10/91.11 & 89.57/89.57 & 87.53/87.54 & 85.26/85.30 & 82.18/82.26 & 80.63/80.72 \\
        \hline
        ResNeXt 101 & 89.26/89.28 & 87.89/87.91 & 87.02/87.04 & 85.60/85.67 & \textbf{83.71/84.83} & 82.76/82.61 \\
        \hline
        ViT b16 & 88.09/88.11 & 86.96/87.03 & 85.91/85.99 & 84.33/84.45 & 82.80/82.69 & 79.48/79.69 \\
        \hline
        ViT b32 & 81.25/81.24 & 81.23/81.22 & 81.12/81.07 & 81.15/81.11 & 80.73/80.68 & 80.14/80.08 \\
        \hline
    \end{tabular}
    \end{adjustbox}
    \caption{Percentage values of Accuracy and F1 Score obtained during the testing phase in three-class classification (GAN vs. DM vs. real) at backbone variation.}
    \vspace{-0.8cm}
    \label{tab:three-inner_testing}
\end{table}

To gain a better understanding of the model's ability to distinguish between real and AI-generated images (from GAN or DM) we recalculated the previous performance values in binary classification: the calculation was performed considering the predicted classes GAN and DM as deepfakes and keeping the predictions of the real class unchanged, then the metrics were recalculated. Table~\ref{tab:binary-inner_testing} shows the metrics obtained from the recalculation. Looking at the new values, we can see how performance has increased in terms of accuracy in both the inference test and, above all, the JPEG compression robustness test. 
%Looking at both tables, we can conclude that the use of DenseNet 161 as the backbone of the Base Model leads to the best classification results and demonstrates good robustness to JPEG compression, 
From the obtained results, DenseNet 161 represents the backbone of the Basic Model as it leads to the best classification results and demonstrates good robustness to JPEG compression: despite the fact that the model was trained using only raw images, the accuracy and F1 score values tend not to decrease drastically as the compression QF decreases.
% Based on these observations, we took DenseNet 161 as the backbone of the Base Model of our final model.}

\begin{table}[t!]
    \centering
    \begin{adjustbox}{max width=1\textwidth}
    \setlength{\tabcolsep}{4pt}
    \begin{tabular}{|c|cccccc|}
        \hline
        \multirow{3}{*}{\begin{tabular}[c]{@{}c@{}}\texttt{Base Model} \\ \texttt{backbone}\end{tabular}} & \multicolumn{6}{c|}{\texttt{Binary  - 
 Accuracy / F1-Score ($\%$)}} \\
         & \multirow{2}{*}{\textit{RAW}} & \multicolumn{5}{c|}{\textit{JPEG-Compression}} \\ \cline{3-7}
         &  & QF90 & QF80 & QF70 & QF60 & QF50 \\
        \hline
        DenseNet 121 & 92.50/92.23 & 91.64/91.35 & \textbf{90.67/90.48} & 87.44/88.55 & 85.86/86.44 & 83.85/84.89 \\
        \hline
        DenseNet 161 & \textbf{93.83/93.67} & \textbf{92.74/92.56} & 90.17/90.16 & \textbf{88.34/88.84} & 82.75/84.28 & 79.65/82.15 \\
        \hline
        DenseNet 169 & 92.10/91.92 & 90.95/90.69 & 89.01/88.92 & 86.21/86.58 & 82.64/83.79 & 80.48/82.12 \\
        \hline
        DenseNet 201 & 93.30/93.03 & 91.30/90.97 & 89.45/89.06 & 88.09/87.81 & 85.78/85.75 & 83.78/83.89 \\
        \hline
        EfficientNet b0 & 89.36/89.01 & 87.72/87.35 & 85.86/85.48 & 84.96/84.57 & 82.69/82.59 & 81.57/81.39 \\
        \hline
        EfficientNet b4 & 86.67/86.56 & 86.03/86.11 & 84.58/84.91 & 84.32/84.67 & 82.14/82.73 & 82.26/82.74 \\
        \hline
        ResNet 18 & 85.62/84.89 & 85.71/84.96 & 84.35/83.49 & 84.05/82.99 & 81.92/80.84 & 81.55/80.21 \\
        \hline
        ResNet 34 & 87.00/86.37 & 86.99/86.28 & 85.25/84.28 & 83.56/82.16 & 81.58/79.89 & 79.57/77.48 \\
        \hline
        ResNet 50 & 90.27/89.83 & 89.10/88.53 & 87.12/86.41 & 85.48/84.71 & 82.92/82.21 & 80.24/79.24 \\
        \hline
        ResNet 101 & 91.59/91.37 & 91.01/90.68 & 89.10/88.60 & 87.72/87.17 & 85.33/84.89 & 84.02/83.63 \\
        \hline
        ResNet 152 & 91.71/91.38 & 90.44/89.93 & 88.57/88.03 & 86.79/86.21 & 84.60/84.16 & 82.72/82.20 \\
        \hline
        ResNeXt 101 & 90.86/90.67 & 90.50/90.25 & 89.12/88.93 & 88.09/87.93 & \textbf{86.54/86.61} & \textbf{86.11/86.13} \\
        \hline
        ViT b16 & 89.49/89.27 & 89.75/89.68 & 88.40/88.43 & 87.19/87.38 & 85.36/85.71 & 82.25/82.59 \\
        \hline
        ViT b32 & 82.31/81.15 & 82.63/81.45 & 82.55/81.38 & 82.24/81.09 & 82.33/80.96 & 81.66/80.20 \\
        \hline
    \end{tabular}
    \end{adjustbox}
    \caption{Percentage values of Accuracy and F1 Score obtained during the testing phase in binary classification (Deepfake vs real) of the model when the backbone varies. Values were obtained considering DM and GAN predictions as Deepfake.}
    \vspace{-0.8cm}
    \label{tab:binary-inner_testing}
\end{table}

\subsection{Comparison with S.O.T.A. in generalization}
In this section, we examine the generalization capacity of our approach. The selected final model uses DenseNet 121 as the backbone of the Base Model, chosen for its excellent performance found in the tests described in Section~\ref{sub:inf_rob}.

Initially, we conducted an analysis of the baselines: the models used as the backbone of the Base Models were trained in the same conditions of our method and subsequently evaluated in terms of generalization. This process allowed us to compare the effectiveness of our model with the use of standard architectures. Next, we compared our model with state-of-the-art models trained on similar tasks, namely the distinction between AI-generated images and real images.

In order to assess the generalisation capability of the models, we used different test sets. These test sets were divided into two categories: images generated by generative models previously observed during the training phase, but with different semantic variations and initial conditions, factors that often complicate classification, and images generated by models not included in the training phase. In addition, we conducted further tests distinguishing between images generated exclusively by GANs technologies, images generated exclusively by DMs technologies and images generated by both technologies. 
We use the notation whereby we define: $\mathcal{T}_*^{i}$ the dataset containing images generated by models already considered in the training phase; $\mathcal{T}_*^{o}$ the dataset containing images generated by architectures not considered in the training phase; $\mathcal{T}_*^{i/o}$ contains images generated by both type of architectures during the training phase; $\mathcal{T}_G^{*}$ the dataset containing only images generated by GANs as fakes; $\mathcal{T}_D^{*}$ the dataset containing only images generated by DMs as fakes; $\mathcal{T}_{D/G}^{*}$ contains images generated by both GANs and DMs architectures. Explicitly:
\begin{itemize}
    \item $\mathcal{T}_G^{i}$ contains a fake image sample of 2000 divided equally between images generated by GauGAN~\cite{park2019gaugan}, BigGAN~\cite{brock2018large}, ProGAN~\cite{karras2017progressive}, and CycleGAN~\cite{zhu2017unpaired}.
    \item $\mathcal{T}_G^{o}$ contains a fake image sample of 2000 divided equally between images generated by Generative Adversarial Transformers (GANformer)~\cite{hudson2021generative}, Denoising DiffusionGANs~\cite{xiao2021tackling}, DiffusionGANs~\cite{wang2022diffusion}, ProjectedGANs~\cite{sauer2021projected}, and Taming Transformers~\cite{esser2021taming}.
    \item $\mathcal{T}_G^{i/o}$ contains a fake image sample of 2000 divided equally between images generated by the same generative models of $\mathcal{T}_G^{i}$ and $\mathcal{T}_G^{o}$.
    \item $\mathcal{T}_D^{i}$ contains a fake image sample of 2000 divided equally between images generated by Diffusion and images taken randomly from the COCOFake dataset~\cite{cocchi2023unveiling}, generated by Stable Diffusion~\footnote{\href{https://github.com/CompVis/stable-diffusion}{github.com/CompVis/stable-diffusion}}.
    \item $\mathcal{T}_D^{o}$ contains a fake image sample of 2000 divided equally between images generated by Vector Quantized Diffusion Model (VQ Diffusion)~\cite{gu2022vector}, Denoising Diffusion Probabilistic Model (DDPM)~\cite{ho2020denoising}, and images taken randomly from the COCOGlide dataset, generated by Glide~\cite{nichol2021glide}.
    \item $\mathcal{T}_D^{i/o}$ contains a fake image sample of 2000 divided equally between images generated by the same generative models of $\mathcal{T}_D^{i}$ and $\mathcal{T}_D^{o}$.
    \item $\mathcal{T}_{D/G}^{i}$ contains a fake image sample of 2000 divided equally between images generated by the same generative models of $\mathcal{T}_D^{i}$ and $\mathcal{T}_G^{i}$.
    \item $\mathcal{T}_{D/G}^{o}$ contains a fake image sample of 2000 divided equally between images generated by the same generative models of $\mathcal{T}_D^{o}$ and $\mathcal{T}_G^{o}$.
    \item $\mathcal{T}_{D/G}^{i/o}$ contains a fake image sample of 2000 divided equally between images generated by all the same previous generative models.
\end{itemize}

We also specify that each of the datasets listed above contains a sample of 2000 real images taken randomly in equal numbers from the datasets We also specify that each of the datasets listed above contains a sample of 2000 real images taken randomly in equal numbers from the AFHQ~\cite{choi2020stargan}, Imagenet~\cite{5206848} and COCO~\cite{lin2014microsoft} datasets.\\
Table~\ref{tab:test_sota} shows the percentage values of the accuracies obtained by the various models in the different contexts $\mathcal{T}$. When reading the results, it is important to consider that all images in the test sets are compressed in JPEG format, which, taking into account that our model was trained using only raw images, may have lowered its performance as demonstrated in Section~\ref{sub:inf_rob}.
The state-of-the-art approaches used for comparison are~\cite{arshed2024multiclass,gandhi2020adversarial,guarnera2024mastering,wang2020cnn}.
This choice is due to the fact that almost all these methods were trained using generative architectures considered in our experiments. 
Wang et al.~\cite{wang2020cnn} and Gandhi et al.~\cite{gandhi2020adversarial} used only images generated by GAN models and represent some of the best approaches in literature able to solve well the deepfake detection task (in the specific domain of GAN generated images). Despite this, experimental results reported in Table~\ref{tab:test_sota} show that these approaches are able to achieve similar classification results compared to methods trained considering images generated by also DM engines. However, these results show little ability to generalize. Our approach is able to generalize better, outperforming such state-of-the-art methods with classification accuracy over 10\%, in any context. Arshed et al.~\cite{arshed2024multiclass} and Guarnera et al.~\cite{guarnera2024mastering} used one specific architecture to extract features for images generated by GAN and DM engines. The main limitation compared to our approach regards the strategy for feature extraction, since we used three specific models to better extract the most discriminative characteristics of the input data for each involved image category (GAN-generated, DM-generated, real).\\
%Except for~\cite{wang2020cnn}, which used only images generated by GAN models, all the others used images synthesised by both technologies.
\begin{table}[t!]
    \centering
    \begin{adjustbox}{max width=1\textwidth}
    \renewcommand{\arraystretch}{1.3}
    \setlength{\tabcolsep}{4pt}
    \begin{tabular}{ccccccccccc}
        \hline
         &  & $\mathcal{T}_G^{i}$ & $\mathcal{T}_G^{o}$ & $\mathcal{T}_G^{i/o}$ & $\mathcal{T}_D^{i}$ & $\mathcal{T}_D^{o}$ & $\mathcal{T}_D^{i/o}$ & $\mathcal{T}_{G/D}^{i}$ & $\mathcal{T}_{G/D}^{o}$ &  $\mathcal{T}_{G/D}^{i/o}$ \\
         \hline
         \hline
         \multirow{14}{*}{\rotatebox{90}{\small{Baselines}}}& DenseNet 121  & 56.57 & \large{\textbf{74.02}} & 66.96 & 72.07 & 48.20 & 58.68 & 60.58 & 64.23 & 63.13 \\
         & DenseNet 161 & 53.56 & 73.97 & 66.21 & 69.19 & 48.12 & 56.93 & 58.55 & 64.51 & 61.98 \\
         & DenseNet 169 & 52.03 & 66.73 & 61.10 & 65.25 & 43.98 & 52.61 & 56.80 & 57.93 & 57.51 \\
         & DenseNet 201 & 55.93 & 70.03 & 64.14 & 67.39 & 48.70 & 56.10 & 60.71 & 62.35 & 62.72 \\
         & EfficientNet b0 & 49.58 & 71.81 & 62.99 & 69.22 & 45.12 & 55.19 & 55.78 & 61.67 & 59.85 \\
         & EfficientNet b4 & 50.37 & 68.64 & 61.36 & 69.42 & 46.60 & 55.92 & 56.75 & 60.98 & 60.00 \\
         & ResNet 18 & 63.77 & 68.89 & 66.65 & 66.23 & 55.03 & 58.30 & 63.70 & 62.30 & 62.82 \\
         & ResNet 34 & 53.87 & 70.03 & 63.25 & 65.48 & 48.55 & 54.76 & 57.31 & 61.84 & 60.98 \\
         & ResNet 50 & 59.58 & 73.13 & 67.95 & 67.89 & 53.38 & 58.50 & 62.10 & 65.01 & 63.90 \\
         & ResNet 101 & 60.35 & 68.08 & 65.40 & 72.12 & 56.45 & 59.68 & 64.21 & 63.62 & 63.20 \\
         & ResNet 152 & 53.94 & 68.61 & 61.84 & 63.90 & 50.15 & 55.27 & 55.27 & 61.51 & 60.00 \\
         & ResNeXt 101 & 54.35 & 67.42 & 62.86 & \large{\textbf{74.18}} & 50.23 & 59.57 & 61.19 & 61.06 & 61.87 \\
         & ViT b16 & 65.81 & 73.31 & 69.46 & 68.59 & 52.69 & 58.30 & \large{\textbf{66.62}} & 64.53 & 62.72 \\
         & ViT b32 & 54.07 & 61.91 & 58.87 & 60.34 & 41.87 & 47.92 & 56.22 & 54.25 & 57.38 \\
         \hline
         \hline
         \multirow{4}{*}{\rotatebox{90}{\small{SOTA}}}&  Gandhi2020~\cite{gandhi2020adversarial} & 52.30 & 50.79 & 51.71 & 49.91 & 50.86 & 50.34 & 51.54 & 50.57 & 51.06 \\
         & Wang2020~\cite{wang2020cnn} & 62.41 & 53.18 & 57.87 & 50.13 & 50.93 & 50.44 & 58.26 & 52.14 & 54.86 \\
          & Arshed2024~\cite{arshed2024multiclass} & 47.46 & 47.65 & 48.54 & 52.69 & 50.00 & 51.04 & 49.89 & 48.94 & 52.20 \\
         & Guarnera2024~\cite{guarnera2024mastering} & 55.00 & 55.63 & 56.23 & 54.11 & 45.98 & 49.97 & 56.07 & 52.21 & 57.17 \\
         \hline
         \hline
         & \textbf{Our} & \large{\textbf{64.74}} & 72.47 & \large{\textbf{69.89}} & 68.09 & \large{\textbf{60.82}} & \large{\textbf{59.96}} & 66.06 & \large{\textbf{65.02}} & \large{\textbf{64.39}} \\
         \hline
    \end{tabular}
    \end{adjustbox}
    \caption{Percentage values of the accuracy obtained in generalization phase. The tests distinguished between images generated from architectures seen in the training phase, but with different initial conditions (superscript i), and images generated from architectures never seen before (superscript o), and mixed (superscript i/o). Furthermore, the tests distinguished between using only images generated by GANs (G-index), those by DMs (D-index), and mixed (G/D-index).}
    \label{tab:test_sota}
    \vspace{-1cm}
\end{table}
In summary, from the obtained results (Table \ref{tab:test_sota}), our approach succeeds on average in generalizing better in most of the performed tests. Although baselines perform well in generalization when the dataset is composed of deepfake images generated by a single technology, they encounter difficulties when the dataset contains images from multiple generating architectures, both seen and unseen (column $\mathcal{T}_{G/D}^{i/o}$). 
Then, the proposed model outperforms all other state-of-the-art methods, confirming the good generalization ability in different contexts.
% This confirms the effectiveness of our block-based approach and prompts us to search for new techniques to extract maximum information from the extracted features in order to solve the generalisation task in the field of Deepfake Detection.  
\label{sec:results}

\section{Conclusion and future works}
The challenge of generalization emerges as a major obstacle in the context of deepfake detection. The ability to accurately distinguish between AI-generated and real images is crucial to monitor the ongoing development of generative models. In this article we proposed a new approach that can ensure robustness to JPEG attacks, typically used by social networks, and contributed to a small step forward in solving the problem of generalization of detectors. The use of three different blocks specialized in the extraction of discriminative features of a specific images category (GAN-generated, DM-generated, real) allows our approach to develop a deeper understanding of intrinsic characteristic between real and synthetic images. This approach aims to provide a solid basis for the accurate identification of images even in the presence of variations and complexity introduced by different image generation techniques. This is the starting point for our future research: we want to strengthen the capabilities of the three discriminative feature extractors, analyze their outputs spatially and model new high-performance and structure-independent architectures.
\label{sec:conclusion}

\subsubsection{Acknowledgements} Orazio Pontorno is a PhD candidate enrolled in the National PhD in Artificial Intelligence, XXXIX cycle, course on Health and life sciences, organized by Università Campus Bio-Medico di Roma.
This research is supported by Azione IV.4 - ``Dottorati e contratti di ricerca su  tematiche dell’innovazione" del nuovo Asse IV del PON Ricerca e Innovazione 2014-2020 “Istruzione e ricerca  per il recupero - REACT-EU”- CUP: E65F21002580005.

\bibliographystyle{splncs04.bst}
\bibliography{references}

\begin{thebibliography}{10}
\providecommand{\url}[1]{\texttt{#1}}
\providecommand{\urlprefix}{URL }
\providecommand{\doi}[1]{https://doi.org/#1}

\bibitem{arshed2024multiclass}
Arshed, M.A., Mumtaz, S., Ibrahim, M., Dewi, C., Tanveer, M., Ahmed, S.: {Multiclass AI-Generated Deepfake Face Detection Using Patch-Wise Deep Learning Model}. Computers  \textbf{13}(1), ~31 (2024)

\bibitem{asnani2023reverse}
Asnani, V., Yin, X., Hassner, T., Liu, X.: {Reverse Engineering of Generative Models: Inferring Model Hyperparameters from Generated Images}. IEEE Transactions on Pattern Analysis and Machine Intelligence  (2023)

\bibitem{bergmann2024forensic}
Bergmann, S., Moussa, D., Brand, F., Kaup, A., Riess, C.: {Forensic analysis of AI-compression traces in spatial and frequency domain}. Pattern Recognition Letters  (2024)

\bibitem{brock2018large}
Brock, A., Donahue, J., Simonyan, K.: {Large Scale GAN Training for High Fidelity Natural Image Synthesis}. In: International Conference on Learning Representations (2018)

\bibitem{cho2019image}
Cho, W., Choi, S., Park, D.K., Shin, I., Choo, J.: {Image-To-Image Translation via Group-Wise Deep Whitening-and-Coloring Transformation}. In: Proceedings of the IEEE/CVF Conference on Computer Vision and Pattern Recognition. pp. 10639--10647 (2019)

\bibitem{choi2018stargan}
Choi, Y., Choi, M., Kim, M., Ha, J.W., Kim, S., Choo, J.: {StarGAN: Unified Generative Adversarial Networks for Multi-Domain Image-to-Image Translation}. In: Proceedings of the IEEE Conference on Computer Vision and Pattern Recognition. pp. 8789--8797 (2018)

\bibitem{choi2020stargan}
Choi, Y., Uh, Y., Yoo, J., Ha, J.W.: {StarGAN v2: Diverse Image Synthesis for Multiple Domains}. In: Proceedings of the IEEE/CVF Conference on Computer Vision and Pattern Recognition. pp. 8188--8197 (2020)

\bibitem{cocchi2023unveiling}
Cocchi, F., Baraldi, L., Poppi, S., Cornia, M., Cucchiara, R.: {Unveiling the Impact of Image Transformations on Deepfake Detection: An Experimental Analysis}. In: International Conference on Image Analysis and Processing. pp. 345--356. Springer (2023)

\bibitem{concas2022tensor}
Concas, S., Perelli, G., Marcialis, G.L., Puglisi, G.: {Tensor-Based Deepfake Detection In Scaled And Compressed Images}. In: 2022 IEEE International Conference on Image Processing (ICIP). pp. 3121--3125. IEEE (2022)

\bibitem{corvi2023detection}
Corvi, R., Cozzolino, D., Zingarini, G., Poggi, G., Nagano, K., Verdoliva, L.: {On the Detection of Synthetic Images Generated by Diffusion Models}. In: IEEE International Conference on Acoustics, Speech and Signal Processing (ICASSP). pp.~1--5. IEEE (2023)

\bibitem{corvi2023intriguing}
Corvi, R., Cozzolino, D., Poggi, G., Nagano, K., Verdoliva, L.: {Intriguing Properties of Synthetic Images: from Generative Adversarial Networks to Diffusion Models}. In: Proceedings of the IEEE/CVF Conference on Computer Vision and Pattern Recognition. pp. 973--982 (2023)

\bibitem{5206848}
Deng, J., Dong, W., Socher, R., Li, L.J., Li, K., Fei-Fei, L.: Imagenet: A large-scale hierarchical image database. In: 2009 IEEE Conference on Computer Vision and Pattern Recognition. pp. 248--255 (2009). \doi{10.1109/CVPR.2009.5206848}

\bibitem{dosovitskiy2020image}
Dosovitskiy, A., Beyer, L., Kolesnikov, A., Weissenborn, D., Zhai, X., Unterthiner, T., Dehghani, M., Minderer, M., Heigold, G., Gelly, S., et~al.: {An Image is Worth 16x16 Words: Transformers for Image Recognition at Scale}. In: International Conference on Learning Representations (2020)

\bibitem{esser2021taming}
Esser, P., Rombach, R., Ommer, B.: {Taming Transformers for High-Resolution Image Synthesis}. In: Proceedings of the IEEE/CVF Conference on Computer Vision and Pattern Recognition. pp. 12873--12883 (2021)

\bibitem{JOEL_ICML_2020}
Frank, J., Eisenhofer, T., Sch{\"{o}}nherr, L., Fischer, A., Kolossa, D., Holz, T.: {Leveraging Frequency Analysis for Deep Fake Image Recognition}. In: Proceedings of the 37th International Conference on Machine Learning, {ICML}. pp. 3247--3258. {PMLR} (2020)

\bibitem{gandhi2020adversarial}
Gandhi, A., Jain, S.: {Adversarial Perturbations Fool Deepfake Detectors}. In: 2020 International Joint Conference on Neural Networks (IJCNN). pp.~1--8. IEEE (2020)

\bibitem{goodfellow2014generative}
Goodfellow, I., Pouget-Abadie, J., Mirza, M., Xu, B., Warde-Farley, D., Ozair, S., Courville, A., Bengio, Y.: {Generative Adversarial Nets}. Advances in Neural Information Processing Systems  \textbf{27} (2014)

\bibitem{gu2022vector}
Gu, S., Chen, D., Bao, J., Wen, F., Zhang, B., Chen, D., Yuan, L., Guo, B.: {Vector Quantized Diffusion Model for Text-to-Image Synthesis}. In: Proceedings of the IEEE/CVF Conference on Computer Vision and Pattern Recognition. pp. 10696--10706 (2022)

\bibitem{guarnera2020fighting}
Guarnera, L., Giudice, O., Battiato, S.: {Fighting Deepfake by Exposing the Convolutional Traces on Images}. IEEE Access  \textbf{8},  165085--165098 (2020)

\bibitem{guarnera2024mastering}
Guarnera, L., Giudice, O., Battiato, S.: {Mastering Deepfake Detection: A Cutting-Edge Approach to Distinguish GAN and Diffusion-Model Images}. ACM Transactions on Multimedia Computing, Communications and Applications  (2024). \doi{10.1145/3652027}

\bibitem{guarnera}
Guarnera, L., Giudice, O., Nastasi, C., Battiato, S.: {Preliminary Forensics Analysis of Deepfake Images}. In: 2020 AEIT International Annual Conference (AEIT). pp.~1--6. IEEE (2020). \doi{10.23919/AEIT50178.2020.9241108}

\bibitem{guarnera2022exploitation}
Guarnera, L., Giudice, O., Nie{\ss}ner, M., Battiato, S.: {On the Exploitation of Deepfake Model Recognition}. In: Proceedings of the IEEE/CVF Conference on Computer Vision and Pattern Recognition. pp. 61--70 (2022)

\bibitem{he2016deep}
He, K., Zhang, X., Ren, S., Sun, J.: {Deep Residual Learning for Image Recognition}. In: Proceedings of the IEEE Conference on Computer Vision and Pattern Recognition. pp. 770--778 (2016)

\bibitem{he2019attgan}
He, Z., Zuo, W., Kan, M., Shan, S., Chen, X.: {AttGAN: Facial Attribute Editing by Only Changing What You Want}. IEEE Transactions on Image Processing (11),  5464--5478 (2019)

\bibitem{ho2020denoising}
Ho, J., Jain, A., Abbeel, P.: {Denoising Diffusion Probabilistic Models}. Advances in Neural Information Processing Systems  \textbf{33},  6840--6851 (2020)

\bibitem{huang2017densely}
Huang, G., Liu, Z., Van Der~Maaten, L., Weinberger, K.Q.: {Densely Connected Convolutional Networks}. In: Proceedings of the IEEE Conference on Computer Vision and Pattern Recognition. pp. 4700--4708 (2017)

\bibitem{hudson2021generative}
Hudson, D.A., Zitnick, L.: {Generative Adversarial Transformers}. In: International Conference on Machine Learning. pp. 4487--4499. PMLR (2021)

\bibitem{karras2018progressive}
Karras, T., Aila, T., Laine, S., Lehtinen, J.: {Progressive Growing of GANs for Improved Quality, Stability, and Variation}. In: International Conference on Learning Representations (ICLR) 2018 (2018)

\bibitem{karras2017progressive}
Karras, T., Aila, T., Laine, S., Lehtinen, J.: {Progressive Growing of GANs for Improved Quality, Stability, and Variation}. In: International Conference on Learning Representations (2018)

\bibitem{karras2021alias}
Karras, T., Aittala, M., Laine, S., H{\"a}rk{\"o}nen, E., Hellsten, J., Lehtinen, J., Aila, T.: {Alias-Free Generative Adversarial Networks}. Advances in Neural Information Processing Systems  \textbf{34},  852--863 (2021)

\bibitem{karras2019style}
Karras, T., Laine, S., Aila, T.: {A Style-Based Generator Architecture for Generative Adversarial Networks}. In: Proceedings of the IEEE/CVF Conference on Computer Vision and Pattern Recognition. pp. 4401--4410 (2019)

\bibitem{karras2020analyzing}
Karras, T., Laine, S., Aittala, M., Hellsten, J., Lehtinen, J., Aila, T.: {Analyzing and Improving the Image Quality of StyleGAN}. In: Proceedings of the IEEE/CVF Conference on Computer Vision and Pattern Recognition. pp. 8110--8119 (2020)

\bibitem{leotta2023not}
Leotta, R., Giudice, O., Guarnera, L., Battiato, S.: {Not with My Name! Inferring Artists’ Names of Input Strings Employed by Diffusion Models}. In: International Conference on Image Analysis and Processing. pp. 364--375. Springer (2023)

\bibitem{lin2014microsoft}
Lin, T.Y., Maire, M., Belongie, S., Hays, J., Perona, P., Ramanan, D., Doll{\'a}r, P., Zitnick, C.L.: {Microsoft Coco: Common Objects in Context}. In: Computer Vision--ECCV 2014: 13th European Conference, Zurich, Switzerland, September 6-12, 2014, Proceedings, Part V 13. pp. 740--755. Springer (2014)

\bibitem{liu2021spatial}
Liu, H., Li, X., Zhou, W., Chen, Y., He, Y., Xue, H., Zhang, W., Yu, N.: {Spatial-Phase Shallow Learning: Rethinking Face Forgery Detection in Frequency Domain}. In: Proceedings of the IEEE/CVF Conference on Computer Vision and Pattern Recognition. pp. 772--781 (2021)

\bibitem{liu2015faceattributes}
Liu, Z., Luo, P., Wang, X., Tang, X.: {Deep Learning Face Attributes in the Wild}. In: Proceedings of International Conference on Computer Vision (ICCV) (December 2015)

\bibitem{Marra2019DoGL}
Marra, F., Gragnaniello, D., Verdoliva, L., Poggi, G.: {Do GANs Leave Artificial Fingerprints?} 2019 IEEE Conference on Multimedia Information Processing and Retrieval (MIPR) pp. 506--511 (2019)

\bibitem{mccloskey2019detecting}
McCloskey, S., Albright, M.: {Detecting GAN-Generated Imagery Using Saturation Cues}. In: 2019 IEEE International Conference on Image Processing (ICIP). pp. 4584--4588. IEEE (2019)

\bibitem{nichol2021glide}
Nichol, A.Q., Dhariwal, P., Ramesh, A., Shyam, P., Mishkin, P., Mcgrew, B., Sutskever, I., Chen, M.: {GLIDE: Towards Photorealistic Image Generation and Editing with Text-Guided Diffusion Models}. In: International Conference on Machine Learning. pp. 16784--16804. PMLR (2022)

\bibitem{park2019gaugan}
Park, T., Liu, M.Y., Wang, T.C., Zhu, J.Y.: {GauGAN: Semantic Image Synthesis with Spatially Adaptive Normalization}. In: ACM SIGGRAPH 2019 Real-Time Live! pp.~1--1 (2019)

\bibitem{ramesh2022hierarchical}
Ramesh, A., Dhariwal, P., Nichol, A., Chu, C., Chen, M.: {Hierarchical Text-Conditional Image Generation with CLIP Latents}. arXiv preprint:2204.06125  \textbf{1}(2), ~3 (2022)

\bibitem{rombach2022high}
Rombach, R., Blattmann, A., Lorenz, D., Esser, P., Ommer, B.: {High-Resolution Image Synthesis with Latent Diffusion Models}. In: Proceedings of the IEEE/CVF Conference on Computer Vision and Pattern Recognition. pp. 10684--10695 (2022)

\bibitem{sauer2021projected}
Sauer, A., Chitta, K., M{\"u}ller, J., Geiger, A.: {Projected GANs Converge Faster}. Advances in Neural Information Processing Systems  \textbf{34},  17480--17492 (2021)

\bibitem{sha2023fake}
Sha, Z., Li, Z., Yu, N., Zhang, Y.: {De-fake: Detection and Attribution of Fake Images Generated by Text-to-Image Generation Models}. In: Proceedings of the 2023 ACM SIGSAC Conference on Computer and Communications Security. pp. 3418--3432 (2023)

\bibitem{shan2023glaze}
Shan, S., Cryan, J., Wenger, E., Zheng, H., Hanocka, R., Zhao, B.Y.: {Glaze: Protecting Artists from Style Mimicry by $\{$Text-to-Image$\}$ Models}. In: 32nd USENIX Security Symposium (USENIX Security 23). pp. 2187--2204 (2023)

\bibitem{sohl2015deep}
Sohl-Dickstein, J., Weiss, E., Maheswaranathan, N., Ganguli, S.: {Deep unsupervised Learning Using Nonequilibrium Thermodynamics}. In: International Conference on Machine Learning. pp. 2256--2265. PMLR (2015)

\bibitem{tan2019efficientnet}
Tan, M., Le, Q.: {Efficientnet: Rethinking Model Scaling for Convolutional Neural Networks}. In: International Conference on Machine Learning. pp. 6105--6114. PMLR (2019)

\bibitem{vyas2023provable}
Vyas, N., Kakade, S.M., Barak, B.: {On Provable Copyright Protection for Generative Models}. In: International Conference on Machine Learning. pp. 35277--35299. PMLR (2023)

\bibitem{wang2021fakespotter}
Wang, R., Juefei-Xu, F., Ma, L., Xie, X., Huang, Y., Wang, J., Liu, Y.: {FakeSpotter: a Simple Yet Robust Baseline for Spotting AI-Synthesized Fake Faces}. In: Proceedings of the Twenty-Ninth International Conference on International Joint Conferences on Artificial Intelligence. pp. 3444--3451 (2021)

\bibitem{wang2020cnn}
Wang, S.Y., Wang, O., Zhang, R., Owens, A., Efros, A.A.: {CNN-Generated Images are Surprisingly Easy to Spot... for Now}. In: Proceedings of the IEEE/CVF Conference on Computer Vision and Pattern Recognition. pp. 8695--8704 (2020)

\bibitem{wang2022diffusion}
Wang, Z., Zheng, H., He, P., Chen, W., Zhou, M.: {Diffusion-GAN: Training GANs with Diffusion}. arXiv preprint arXiv:2206.02262  (2022)

\bibitem{xiao2021tackling}
Xiao, Z., Kreis, K., Vahdat, A.: {Tackling the Generative Learning Trilemma with Denoising Diffusion GANs}. arXiv preprint arXiv:2112.07804  (2021)

\bibitem{xie2017aggregated}
Xie, S., Girshick, R., Doll{\'a}r, P., Tu, Z., He, K.: {Aggregated Residual Transformations for Deep Neural Networks}. In: Proceedings of the IEEE Conference on Computer Vision and Pattern Recognition. pp. 1492--1500 (2017)

\bibitem{zhu2017unpaired}
Zhu, J.Y., Park, T., Isola, P., Efros, A.A.: {Unpaired Image-To-Image Translation Using Cycle-Consistent Adversarial Networks}. In: Proceedings of the IEEE International Conference on Computer Vision. pp. 2223--2232 (2017)

\end{thebibliography}
% \input{references}
% \end{thebibliography}
\end{document}